\documentclass[10pt,journal,compsoc]{IEEEtran}
\usepackage{float}
\usepackage{cite}
\usepackage[numbers,sort&compress]{natbib}
\usepackage{graphicx}
\usepackage{stfloats}
\usepackage{amsmath}
\usepackage{amsfonts,amssymb}

\usepackage{lineno,hyperref}
\usepackage{epstopdf}
\usepackage{mathrsfs}
\usepackage{chapterbib}
\usepackage{algorithm}
\usepackage{algpseudocode}
\usepackage{subfigure}
\usepackage{multirow}
\usepackage{hyperref}
\ifCLASSINFOpdf
 
\else
\fi

\begin{document}

\title{Cascading Hierarchical Networks with Multi-task Balanced Loss for Fine-grained hashing}

\author{Xianxian~Zeng and Yanjun~Zheng%~\IEEEmembership{Member,~IEEE,}
        
        %~Shun~Liu,
        %~Jun~Yuan,
        %~Xiaodong~Wang,
        %and ~Weijun~Yang
        %and~Jane~Doe,~\IEEEmembership{Life~Fellow,~IEEE}% <-this % stops a space
\IEEEcompsocitemizethanks{
	\IEEEcompsocthanksitem Xianxian Zeng is with the Department of Computer Science, Guangdong Polytechnic Normal University, Guangzhou, China, 510000. And he is also an adviser in the Guangdong Provincial Key Laboratory of Big Data Computing, The Chinese University of Hong Kong, Shenzhen. \protect\\
\IEEEcompsocthanksitem Yanjun Zheng is with the Department of Computer Science, Guangdong Polytechnic Normal University, Guangzhou, China, 510000.\protect\\
\IEEEcompsocthanksitem Xianxian Zeng(zengxianxian@gpnu.edu.cn) is the corresponding author for this work. And Yanjun Zheng is the first student author. \protect\\
}% <-this % stops an unwanted space
}

\markboth{Journal of \LaTeX\ Class Files,~Vol.~14, No.~8, August~2015}%
{Shell \MakeLowercase{\textit{et al.}}: Bare Demo of IEEEtran.cls for Computer Society Journals}

\IEEEtitleabstractindextext{%
\begin{abstract}
	With the explosive growth in the number of fine-grained images in the Internet era, it has become a challenging problem to perform fast and efficient retrieval from large-scale fine-grained images. Among the many retrieval methods, hashing methods are widely used due to their high efficiency and small storage space occupation. Fine-grained hashing is more challenging than traditional hashing problems due to the difficulties such as low inter-class variances and high intra-class variances caused by the characteristics of fine-grained images. To improve the retrieval accuracy of fine-grained hashing, we propose a cascaded network to learn compact and highly semantic hash codes, and introduce an attention-guided data augmentation method. We refer to this network as a cascaded hierarchical data augmentation network. We also propose a novel approach to coordinately balance the loss of multi-task learning. We do extensive experiments on some common fine-grained visual classification datasets. The experimental results demonstrate that our proposed  method outperforms several state-of-art hashing methods and can effectively improve the accuracy of fine-grained retrieval. The source code is publicly available: \href{https://github.com/kaiba007/FG-CNET}{https://github.com/kaiba007/FG-CNET}.
\end{abstract}

\begin{IEEEkeywords}
Fine-grained hashing, cascading hierarchical networks, balanced loss
\end{IEEEkeywords}}

\maketitle

\IEEEdisplaynontitleabstractindextext

\IEEEpeerreviewmaketitle

\IEEEraisesectionheading{\section{Introduction}\label{sec:introduction}}

\IEEEPARstart{I}{n} the era of big data, the size of fine-grained data\cite{wei2019rpc,hou2017vegfru} in practical applications such as image retrieval has exploded. Due to the storage efficiency and computational efficiency of binary hash codes, hashing\cite{wang2017survey} has become one of the most popular and effective solutions for large-scale image retrieval tasks. The image retrieval task has long been an important research topic in computer vision, which aims to find similar images from large-scale datasets for query images. The specific approach of image retrieval task is to sort the images in the database according to their relevance to the query image and return them in order. The image retrieval method\cite{lin2016learning} using hashing first maps the high-dimensional image data into compact binary codes, and then searches the near-neighborhood data very quickly. Although the hashing method does not obtain precise neighborhood results, it can save storage space and improve the retrieval speed, which is better adapted to the requirements of large-scale image retrieval. In classical computer image processing research tasks, the target object categories for hashing are usually coarse-grained broad categories such as "dogs", "cars" and "birds". However, the target object categories for fine-grained hashing are different sub-categories under the same broad category, such as "sunflower", "dandelion", "rose", etc. under the flower category. In fact, the fine-grained hashing is more challenging, mainly because of the problem of low inter-class variances and high intra-class variances due to the characteristics of fine-grained images themselves. Specifically, the composition of parts is generally the same among targets of different sub-categories, yet there is a rich diversity among individual parts of the same sub-categories. Therefore, obtaining more discriminative image high-level semantic features and generating compact and efficient hash codes are often the key to the fine-grained hashing.
	
	Early research on hashing was mainly based on algorithms with manual features\cite{xie2015fine}. Such algorithms generally studied various hand-designed feature descriptors based on visual cues representing the image (e.g., color, texture, shape, etc.), and then generated hash codes by a simple linear mapping. Due to the limited descriptive power of manual features, the retrieval accuracy of early hashing algorithms based on manual features was generally not high. In recent years, neural networks have achieved excellent performance in a wide range of application areas, showing comparable or even better accuracy than humans on datasets in the areas of image recognition\cite{he2015delving} and speech recognitionn\cite{xiong2016achieving} in particular. The emergence of deep learning\cite{lecun2015deep} has brought tremendous results improvement for fine-grained hashing\cite{gordo2016deep}. Deep hashing algorithms that combine deep convolutional neural networks with hashing have also become the dominant approach and the main alternative to hand-designed feature engineering. The introduction of a fine-grained retrieval method based on deep hashing can effectively solve the problems of large storage space and low retrieval efficiency caused by large-scale image retrieval tasks. Some current fine-grained hashing methods, although they have obtained good results on some fine-grained datasets, still face the following problems: 1) The high intra-class variance and low inter-class variance due to fine-grained image characteristics make images prone to erroneous retrieval and affect retrieval performance. 2) Most network models have limited feature description performance to generate compact and discriminative hash codes. 3) When hashing tasks are combined with other tasks, they rely solely on manual experience to adjust the parameters between different tasks.

	To address the above problems faced in fine-grained hashing, this paper proposes a unified end-to-end trainable network, and our model contains three modules fine-grained representation learning module, hash code learning module, and a novel multi-task balanced loss function. Specifically, to overcome the problem of small inter-class variation and large intra-class variation in fine-grained data and extract more effective image features, we propose to use a cascaded network to leverage the representational power of the network model to learn compact and highly semantic hash codes, thus improving the accuracy of fine-grained image retrieval. Inspired by fine-grained classification, to make feature extraction more meaningful, we classify the topmost features in the cascaded network. It is also worth exploring how to coordinate the learning between two different tasks, the hashing task and the classification task. In contrast to the usual approach using manual tuning of parameters, we propose a novel approach to coordinate the learning of both. To further extract discriminative fine-grained features, we also introduce attention-guided data augmentation. The attention mechanism allocates more attention to the target regions that need attention and automatically ignores some low-value information. The introduction of attention-guided data augmentation allows the network to focus on the target object to be retrieved and ignore some distractions such as image background. 

	Experimental results on several public datasets show that the proposed approach achieves new state-of-the-art performance on fine-grained hashing. We have also performed exhaustive comparison and ablation experiments on the classical fine-grained image dataset CUB. The following are the main contributions of this paper.
\begin{itemize}
\item We propose a unified end-to-end trainable network that uses a cascaded network to learn compact and highly semantic hash codes.
\item We introduce a novel approach to balance the loss of multi-task learning, avoiding the hassle of manually tuning parameters.
\item We also use attention zooming as a data augmentation method. The attention-based guidance allows the network to give more attention to the target features.
\end{itemize}

	The rest of the paper is organized as follows: Section 2 introduces some related work; Section 3 introduces the method used in the experiment; Section 4 describe the results and analysis of experiments; conclusions and future work are presented in section 5.



\section{Releted Work}
	In this section, we review the related works of fine-grained classification, fine-grained retrieval and fine-grained hashing. 

\subsection{Fine-Grained Classification}
	The current mainstream approaches are mainly studied in three directions: introducing Web image training methods, image attention localization algorithms, and integrated learning methods.
To address the problem of insufficient training data for fine-grained image datasets, there are related studies\cite{xu2015augmenting,krause2016unreasonable,zhang2020web,zhang2020web2} that explore from the perspective of increasing training data, and mitigate the overfitting phenomenon of the network model by introducing a large amount of Web image data for adequate training of the DNN model. To achieve discriminative region localization, early strongly supervised signal-based localization algorithms\cite{zhang2014part,huang2016part} mainly used a large amount of object-level and component-level localization annotation information to train Convolutional Neural Network (CNN), prompting CNN to detect salient regions accurately. In recent years, a large number of new weakly supervised attention localization algorithms have been proposed, which employ only image category annotation and do not rely on object localization annotation, avoiding a large amount of time-consuming and costly manual annotation.

	Integration learning can effectively improve the testing performance of models through the combination of multiple different forms of features classifier outputs, and some work focuses on integration at the feature representation level to obtain better fine-grained feature representations. One of the most representative ones is the bilinear CNN\cite{lin2017bilinear} and its extension work\cite{gao2016compact,yu2018hierarchical}, where the convolutional feature maps extracted by different models are bilinearly pooled and fused to obtain pairwise correlations between different feature channels to obtain better fine-grained high-dimensional feature representations. In addition, Cai\cite{cai2017higher} proposed multi-level convolutional features based on higher-order synthesis, which opened a new perspective for fusing convolutional responses from different layers. These fine-grained image classification algorithms based on integrated learning improve the generalization performance of network models by means of integrated learning, which is another important research direction for fine-grained image analysis tasks.

\subsection{Fine-Grained Retrieval}
	The fine-grained image retrieval task expects the feature distance between different categories of images to be greater than the feature distance between images of the same category, so that images of the same category can be retrieved based on the feature distance. Therefore, the current mainstream approaches often use deep metric learning to train deep models, such as using Contrastive loss\cite{bell2015learning}, Triplet loss\cite{zhang2016embedding,huang2016local}, Multi-Similarity loss\cite{wang2019multi} and hash metric loss\cite{jin2020deep}. These methods of metric learning pass a local objective function based on the sample level. Specifically, in a minimal batch, multiple pairs of positive and negative samples are obtained with the expectation of minimizing the distance between positive samples while maximizing the distance between negative samples, allowing the network model to extract discriminative image features.

	Recently, Extensive research\cite{zheng2018centralized,zheng2019towards,qian2019softtriple,zeng2020fine} have shown that global loss functions based on category centers are more beneficial for obtaining highly discriminative fine-grained image features.Zheng\cite{zheng2018centralized} assumed in his experiments that each category in the fine-grained image training dataset has its own feature center, and it is only necessary to make the features extracted by the network model close to their corresponding category feature centers . Therefore, a triadic loss function based on the global metric space was proposed. Subsequently, Zheng\cite{zheng2019towards} found that the cross-entropy loss function, which is commonly used for the classification task, defines the category centers of the classifier and is very similar to the global loss function, but cannot be used directly for the retrieval task. To eliminate the gap between classification and retrieval tasks, they proposed a normalized scaling layer and used a Schmidt orthogonalization optimization model to further enhance fine-grained image retrieval. While Qian\cite{qian2019softtriple} proposed a global-based soft triplet loss function, which allows the training of metric learning without relying on local triplet sampling, Zeng\cite{zeng2020fine} proposed a segmented cross-entropy loss function to address the overfitting phenomenon of the fine-grained image retrieval task model. Their experiments show that the overfitting phenomenon occurring in the training phase of the model can be effectively mitigated by introducing an appropriate amount of noise in the training phase of the network model.

\begin{figure*}[ht]
	\centering
	\includegraphics[width=7 in]{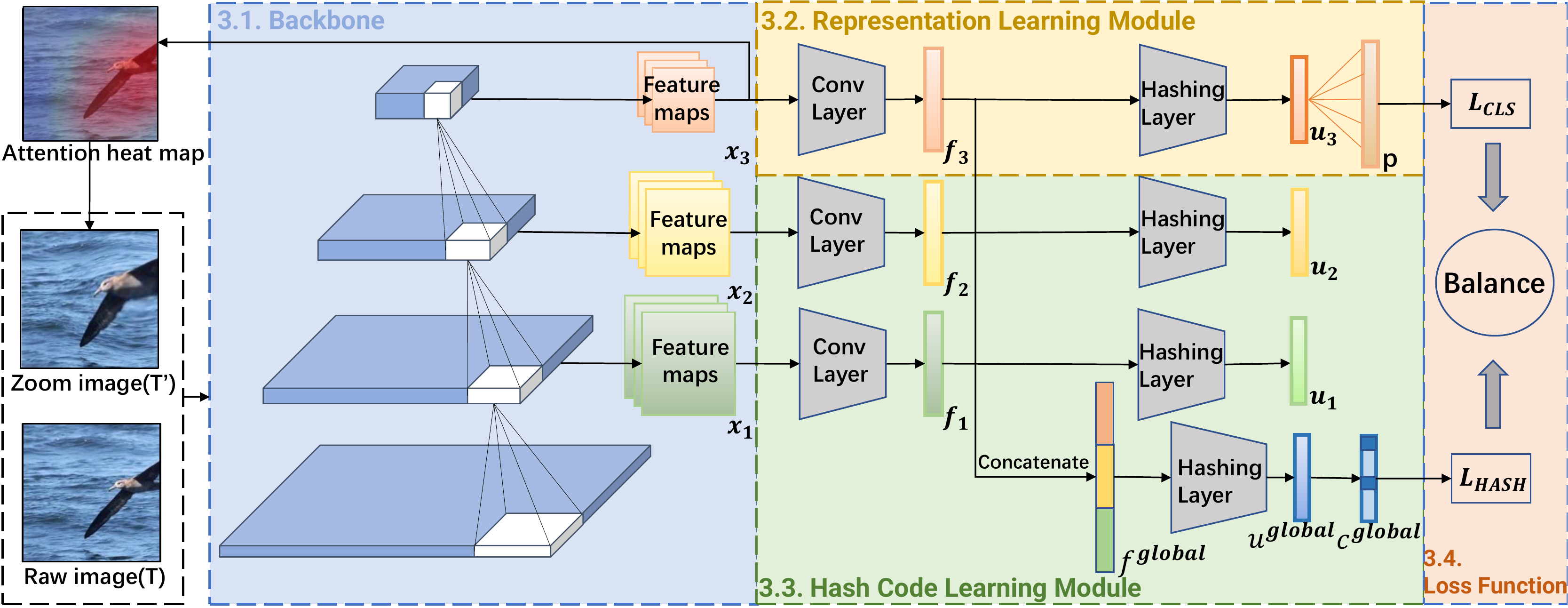}
	\caption{An overview of our proposed framework}
	\label{fig1}
\end{figure*}
\subsection{Fine-Grained Hashing}
	Hashing methods can be divided into data-independent hashing\cite{kulis2009kernelized,datar2004locality} and data-dependent hashing\cite{kong2012isotropic}. Data-independent hashing was the main research method of early researchers. This approach mainly involves manually designed or randomly generated hash functions, and the generated hash functions do not depend on any training data. In order to generate more compact and efficient binary codes, data-dependent hashing methods were proposed. This approach expects to learn hash functions from some training data and is the mainstream approach at present. In recent years, deep hashing methods that combine deep convolutional neural networks with hashing have seen great breakthroughs in performance as neural networks have achieved excellent performance in a wide range of applications.
 
	The deep supervised hash function uses labeled data to map input data points to binary codes. Some deep supervised hashing methods are DPSH\cite{li2015feature}, DSH\cite{liu2016deep}, DSHNP\cite{su2017deep} and ADSH\cite{jiang2018asymmetric}. DPSH\cite{li2015feature} and DSH\cite{liu2016deep} both use CNN models to obtain binary codes from input image pairs and labels indicating whether two images are similar, and design a reasonable loss function to learn hash codes. DSHNP\cite{su2017deep} proposes the idea of implementing a purely nonlinear deep hashing network structure and uses a soft decision tree as a nonlinear mapping function. ADSH\cite{jiang2018asymmetric} uses an asymmetric strategy to learn a hash function. Specifically, it learns the hash function only for query points and learns the database hash code directly, which is different from most deep supervised hashing methods\cite{li2015feature, liu2016deep}, which learn the hash function for both query points and database points.

	In recent years, as fine-grained analysis has become a hot research topic, fine-grained hashing has become a more challenging and practical hashing task. DSaH\cite{jin2020deep} is the first work devoted to fine-grained hashing problem. ExchNet\cite{cui2020exchnet} proposed proposed an end-to-end hash network called ExchNet, which learns hash codes after obtaining discriminative fine-grained representations by using local feature alignment methods. A2-NET\cite{wei20212} proposes an end-to-end Attribute-Aware hashing Network that performs hash code learning by unsupervised attribute-guided learning and attribute-aware features obtained by decorrelating attribute-specific features. To address various problems caused by instance-to-instance loss functions, such as higher training complexity and slower convergence speed, FISH\cite{chen2022fine} designs a proxy-based loss function.

\section{Methodology}
	Fig.~\ref{fig1} shows the proposed  framework. Our method includes four main components: (1)a multi-stage backbone; (2)a fine-grained representation learning module; (3)an attribute-aware hash codes learning module; (4) a loss function[other name-balance]

\subsection{Multi-stage Backbone}
	This subsection introduces the backbone network of our proposed deep hash model. To better extract discriminative image features, we use multi-stage backbone as the Feature extractor for fine-grained images. common multi-stage backbone includes VGG\cite{simonyan2014very}, ResNet\cite{he2016deep} and DenseNet\cite{huang2017densely}, etc. As a powerful feature extraction network, ResNet50 is used as a backbone network for many computer vision tasks, including fine-grained retrieval tasks\cite{li2015feature, cao2017hashnet, jiang2018asymmetric, cui2020exchnet, wei20212, shen2022semicon}. We use a typical pre-trained ResNet50 as the backbone. as shown in Fig.~\ref{fig1} , we denote ResNet50 as $N(\cdot;\theta)$. To prepare for the learning of image representation and hash codes later, we input the original image $T\in \mathbb{R}^{c\times h \times w}$ into ResNet50 and obtain feature maps $X=\{x_1;x_2;……;x_j\}$ from the backbone network of different depths. The formula is shown as follows.
\begin{equation}
\label{E1}
\begin{aligned}
& x_1=N_1(T,\theta_1),\\
& x_2=N_2(x_1,\theta_2),\\
& ……\\
& x_j=N_j(x_{j-1},\theta_j).
\end{aligned}
\end{equation}
where $j \in \{1,... , n \}$ represent different stages and j is proportional to the depth of the backbone network.

\subsection{Representation Learning Module}
\begin{figure*}[htbp]
	\centering
	\includegraphics[width=7 in]{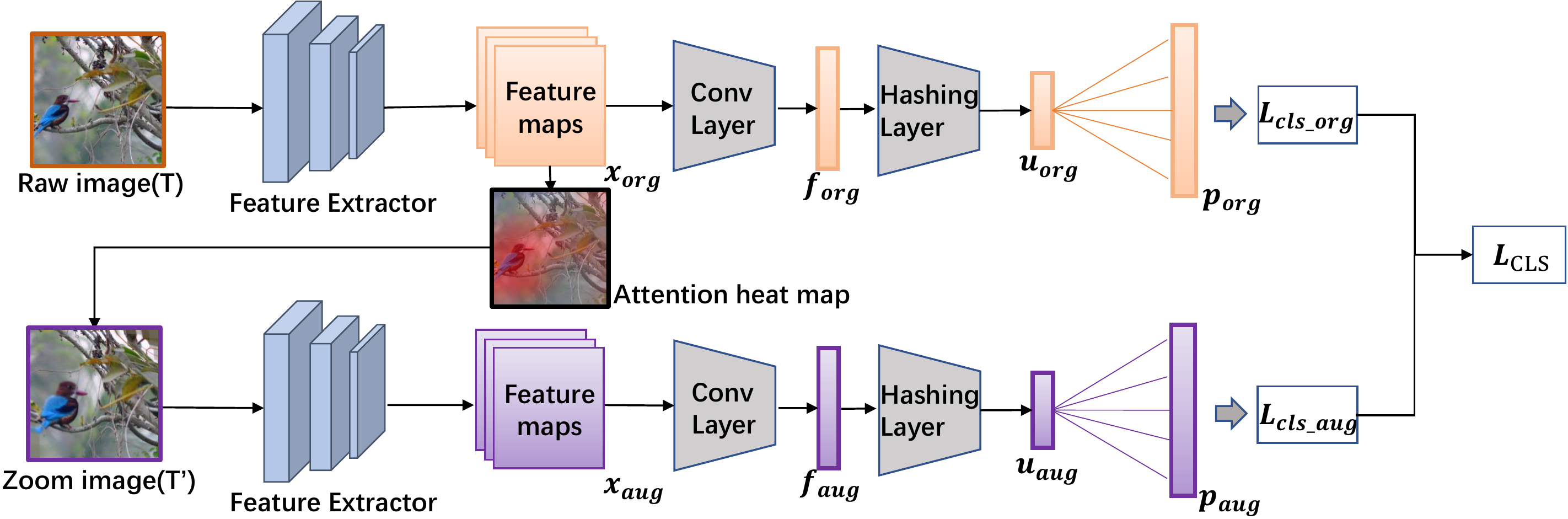}
	\caption{Representation Learning Module}
	\label{fig2}
\end{figure*}
	This subsection introduces the representation learning module for fine-grained images. Inspired by fine-grained classification, we classify the features at the top layer in the cascade network in order to make feature extraction more meaningful. The model is shown in Fig.~\ref{fig2}. As can be seen, our representation learning module contains two main parts: the raw image representation learning part and the attention-guided augmented image representation learning part.

\subsubsection{Raw image representation learning}
	ResNet50 mentioned in Section 3.1 is used as the base model of the fine-grained representation learning module . Specifically, we first input the original image $T$ into the backbone network ResNet50 and take the output of the last layer of ResNet50 to obtain the feature map $X_{org}$. Then, to further extract the features, $X_{org}$ is convolved to obtain the feature $f_{org}$, and the feature $f_{org}$ is mapped through the hash layer to obtain the hash code $c_{org}$ corresponding to the original image. Finally the hash code $c_{org}$ is mapped linearly to obtain the final feature $p_{org}$ used for classification.
\subsubsection{Augmented image representation learning}
	In order to give more attention to the distinguished fine-grained feature regions, we apply an attention mechanism in the images and add attention-guided data augmentation. The data augmentation here follows the attentional zooming approach of paper a. Specifically, We first input the original image T into the network to get the feature map $X_{org}$, then we superimpose $X_{org}$ and use bilinear upsampling to get the attention map $A$. the input image $A$ is normalized to image $A^*$ before attentional zooming, 
\begin{equation}
\label{E2}
A^*=\frac{A}{max(A)}
\end{equation}
Then a Gaussian function is constructed centered at the maximum of the image response, which is unevenly sampled by the system to obtain the attention zooming image $T'$. Finally, the attention-guided enhanced image $T'$ is put into the network and the features $p_{aug}$ used for classification are obtained by the same steps as in section. The details of attention zoom is referred from Saliency-Sampler\cite{recasens2018learning}.
\subsection{Hash Code Learning Module}
	In the hash code learning module, we use a cascaded network to learn hash codes. The backbone network is ResNet50 that shares parameters with the fine-grained representation learning module . As shown in Fig.~\ref{fig2}, first, in Section 3.1 we obtain several intermediate feature maps $X=\{x_1;x_2;……;x_j\}$ of the backbone network with different depths, and feed them into different convolutional blocks for further feature extraction. These convolution blocks are denoted as $C_j(\cdot)$ , where $j \in \{1,... , n \}$ represent different stages and j is proportional to the depth of the subnetwork. The intermediate feature maps $X=\{x_1;x_2;……;x_j\}$ of different depths are passed through the corresponding convolution blocks $C_j(\cdot)$ to obtain different hash features$f_j$.
\begin{equation}
\label{E3}
\begin{aligned}
f_1=C_1(x_1),\\
f_2=C_2(x_2),\\
  ……\\
f_j=C_j(x_j)
\end{aligned}
\end{equation}
Then, the different hash features $f_j$ are followed by a hash layer $H_j(\cdot)$. Therefore, different hash codes $c_j$ can be finally obtained for different depths of the backbone network.

	In order to get the cascaded features with global meaning, we first concatenate $f_j$ and get $f^{global}$. 
\begin{equation}
f^{global}=concat(f_1,f_2,……,f_j)
\end{equation}
Then a linear hash layer is added after this global hash feature $f^{global}$, and the final globally meaningful hash code $c^{global}$ is obtained.
\begin{equation}
c^{global}=H'(f^{global})=sign({W^h}f^{global})
\end{equation}

\subsection{Loss Functions}
As show in Fig. \ref{fig1}, we employ the classification loss $L_{cls}$ and the hashing loss $L_{HASH}$ for the multi-task balanced loss to train the proposed cascading hierarchical network.  

\subsubsection{Classification Loss}
	The loss function for classification uses the cross-entropy loss function that is common in classification tasks. In order to obtain higher accuracy, we transform the label from the original one-hot type to a smoother type by using label smoothing. The specific formula is as follows.
\begin{equation}
\label{sm_y}
\begin{aligned}
y_{i}^{LS}=y_i(1-\lambda)+\lambda/l,\\
s.t. y_i\in\{0,1\},\\
\end{aligned}
\end{equation}
where $l$ is the number of classes in the dataset.

In the network model we performed the classification task for both the original and the augmented images with the following equations.
\begin{equation}
\label{E3}
L_{cls\_org}=-\sum^n_{i=1}y_i^{LS} \cdot log\frac{exp(\overline{y}_i)}{\sum^l_{k=1}exp(\overline{y}_{ik})}
\end{equation}
\begin{equation}
\label{E4}
L_{cls\_aug}=-\sum^n_{i=1}y_i^{LS} \cdot log\frac{exp(\overline{y}'_i)}{\sum^l_{k=1}exp(\overline{y}'_{ik})}
\end{equation}

	The general equation of the classification task is as follows:
\begin{equation}
\label{E8}
L_{CLS}=(L_{cls\_org}+L_{cls\_aug})/2
\end{equation}

\subsubsection{Hash Loss}
It is noted in FISH\cite{chen2022fine} that the proxy-based loss function is more suitable for fine-grained hashing than the instance-based loss function, so this work uses the same proxy-based loss function as FISH as the loss function for hash code learning. Specifically, because of the binary constraint, it is difficult to use the general agent-based functions directly for hashing tasks, so FISH proposes the following formula as the loss function for fine-grained hashing. 
\begin{equation}
\label{E6}
L_{hash}=\sum_{j=1}^l\sum_{i=1}^n(Y_{ji}-d_jc_i)
\end{equation}
where $c_i=H'(f_i^{global})$, $d_j \in \mathbb{R}^{1\times k}$ is the proxy-vector of category k, and Y is the semantic label.

	To allow the binary-based hash loss to be optimized by the back-propagation algorithm of the neural network, FISH changes the above equation to:
\begin{equation}
\label{E10}
\begin{aligned}
L'_{hash}=||Y-DC||^2_F+\sum^n_{i=1}||c_i-W^hf_i^{global})||^2_F,\\
s.t. C\in\{-1,1\}^{k \times n},\\
\end{aligned}
\end{equation}
where $D\in \mathbb{R}^{l\times k}$ is all the proxy-vectors combined into one matrix, $C\in\{+1,-1\}^{k\times n}$ is all the hash codes to be optimized combined into one matrix, and F denotes the Frobenius parametrization of the matrix.

	Eq\ref{E10} is divided into two parts. The left half is the hash code learning problem, which can be solved by alternating iterations before the network is trained. The right half is the regression problem, which can be solved together with the classification problem of Eq\ref{E8} by the standard backpropagation algorithm, as described in detail in Section 3.4.3. Therefore, the hash loss function for network training is as follows.
\begin{equation}
\label{E6}
L_{HASH}=\sum^n_{i=1}||c_i-W^hf_i^{global})||^2_F
\end{equation}

	The optimization of Eq\ref{E10} is divided into two steps. The first step is to optimize the left half of Eq\ref{E10} before network training, with the goal of optimizing the following equation.	
\begin{equation}
\label{E_DC}
\begin{aligned}
\min\limits_{C,D}=||Y-DC||^2_F,\\
s.t. C\in\{-1,1\}^{k\times n},\\
\end{aligned}
\end{equation}
For easier optimization, Eq\ref{E_DC} is transformed into the following equation.
\begin{equation}
\label{E6}
\begin{aligned}
\min\limits_{C,D,E,O}||Y-DE||^2_F+\sigma||C-OE||^2_F\\
s.t. C\in\{-1,1\}^{k\times n},OO^T=I\\
\end{aligned}
\end{equation}
where $E\in \mathbb{R}^{r\times r}$ is a real-valued interval state of $C$, $O$ is an orthogonal rotation matrix that makes $E$ and $C$ as similar as possible, and  is a compromise hyperparameter. Then, the variables D, E, O and C can be optimized alternatively in some way, see FISH in particular. The second step is to optimize the hash problem together with the classification problem during network training, where the hash loss function is the right half of Eq\ref{E10}.

\subsubsection{Total Loss }
	In Section 4 for the hash loss function, we split it into two parts and decouple it into hash code learning and mapping network training. After hash code learning is solved in the first step, the next step is to solve the other part of the hash function together with the classification problem by back propagation algorithm. Generally, for the training of the overall network, the total loss function can be defined as equation \ref{E12}.
\begin{equation}
\label{E12}
L_{TOTAL}=L_{HASH} +L_{CLS}
\end{equation}

	In the equation\ref{E12}, $L_{HASH}$ represents the hashing task and $L_{CLS}$ represents the classification task. For the multi-task learning in the training process, it is important to seek some suitable weighting coefficients for the whole model. The general approach relies on the experimenter to manually adjust the weights of each target, which not only tests the researcher's experience very much, but also occupies valuable computing resources with a large number of experiments. To solve such problems more effectively, we propose a novel multi-task learning balanced objective function that can adaptively update the weights of multi-task objectives while the network is being trained. As shown in the Fig.~\ref{fig4}, we hope to dynamically balance the learning of two types of tasks by the network. Specifically, we add learnable parameters to each learning task. The total loss of the proposed network training process is defined in the following form:
\begin{equation}
\label{E6}
L_{TOTAL}=\frac{1}{\alpha^2} L_{HASH}+ \frac{1}{\beta^2}L_{CLS}+log(\alpha+1)+log(\beta+1)
\end{equation}

where $\alpha$ and $\beta$ are learnable hyper-parameters. Note that to ensure that the training is convergent, we add

\begin{figure}[H]
	\centering
	\includegraphics[width=2.5 in]{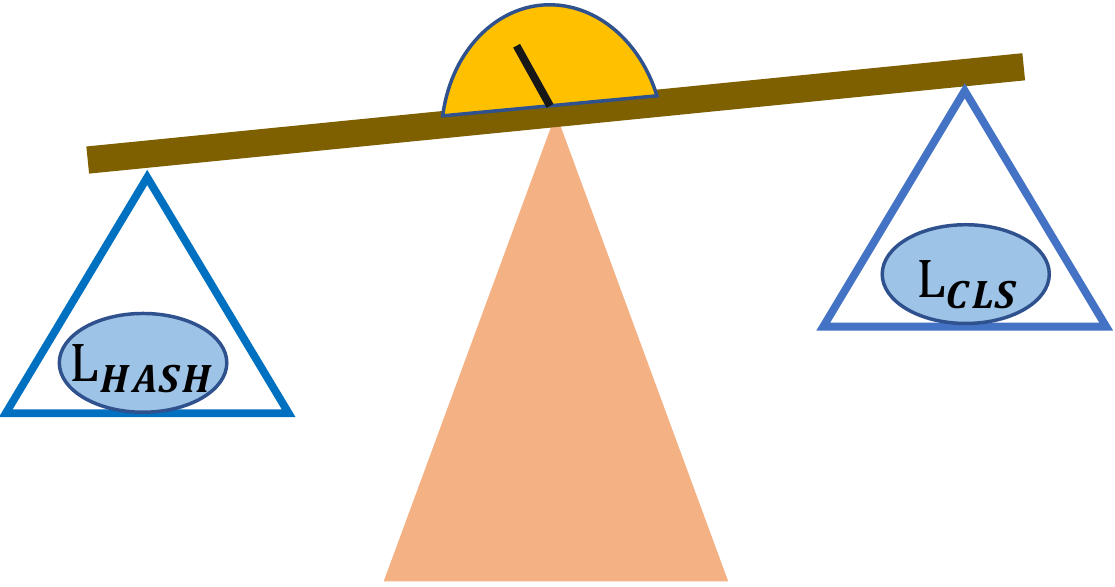}
	\caption{balance}
	\label{fig4}
\end{figure}

%%%%%%%%%%%%%%%%%%%%%%%%%%%%%%%%%%%%%%%%%%%%%%%%%%%%%%%%%%%%%%%%%%%%%%%%%%%%%

\begin{table*}[!t]
	\caption{MAP results for different number of bits on five fine-grained datasets}
	\label{tab1}
	\centering
	\resizebox{17cm}{5cm}{
		\begin{tabular}{|*{9}{c|}}
		\hline
	
		\hline
		\textbf{Datasets}&\textbf{\#bits}&\textbf{DPSH}&\textbf{HashNet}&\textbf{ADSH}&\textbf{ExchNet}				&\textbf{AA-NET}&\textbf{Semicon}&\textbf{OURS}\\
		\hline
			CUB-200-2011&\begin{tabular}[c]{@{}l@{}}12bits\\24bits\\32bits\\48bits\end{tabular}
			&\begin{tabular}[c]{@{}l@{}}0.0868\\0.1251\\0.1274\\0.1558\end{tabular}
			&\begin{tabular}[c]{@{}l@{}}0.1203\\0.1777\\0.1993\\0.2213\end{tabular}
			&\begin{tabular}[c]{@{}l@{}}0.2003\\0.5033\\0.6168\\0.6543\end{tabular} 
			&\begin{tabular}[c]{@{}l@{}}0.2514\\0.5898\\0.6774\\0.7105\end{tabular} 
			&\begin{tabular}[c]{@{}l@{}}0.3383\\0.6101\\0.7161\\0.7733\end{tabular}
			&\begin{tabular}[c]{@{}l@{}}0.3776\\0.6541\\0.7261\\0.7967\end{tabular}
			&\begin{tabular}[c]{@{}l@{}}\textbf{0.7710}\\\textbf{0.8211}\\\textbf{0.8309}\\\textbf{0.8392}\end{tabular}\\
		\hline
			Aircraft&\begin{tabular}[c]{@{}l@{}}12bits\\24bits\\32bits\\48bits\end{tabular}
			&\begin{tabular}[c]{@{}l@{}}0.0874\\0.1087\\0.1354\\0.1394\end{tabular}
			&\begin{tabular}[c]{@{}l@{}}0.1491\\0.1775\\0.1942\\0.2032\end{tabular}         
			&\begin{tabular}[c]{@{}l@{}}0.1554\\0.2309\\0.3037\\0.5065\end{tabular}
			&\begin{tabular}[c]{@{}l@{}}0.3327\\0.4583\\0.5183\\0.5905\end{tabular}
			&\begin{tabular}[c]{@{}l@{}}0.4272\\0.6366\\0.7251\\0.8137\end{tabular}
			&\begin{tabular}[c]{@{}l@{}}0.4987\\0.7508\\0.8045\\0.8423\end{tabular}
			&\begin{tabular}[c]{@{}l@{}}\textbf{0.8615}\\\textbf{0.8827}\\\textbf{0.8840}\\\textbf{0.8917}\end{tabular}\\                       
		\hline
			VegFru&\begin{tabular}[c]{@{}l@{}}12bits\\24bits\\32bits\\48bits\end{tabular}
			&\begin{tabular}[c]{@{}l@{}}0.0633\\0.0905\\0.1028\\0.0911\end{tabular} 
			&\begin{tabular}[c]{@{}l@{}}0.0370\\0.0624\\0.0783\\0.1029\end{tabular}
			&\begin{tabular}[c]{@{}l@{}}0.0824\\0.2490\\0.3653\\0.5515\end{tabular} 
			&\begin{tabular}[c]{@{}l@{}}0.2355\\0.3593\\0.4827\\0.6930\end{tabular}
			&\begin{tabular}[c]{@{}l@{}}0.2552\\0.4473\\0.5275\\0.6977\end{tabular}
			&\begin{tabular}[c]{@{}l@{}}0.3032\\0.5845\\0.6992\\0.7977\end{tabular}
			&\begin{tabular}[c]{@{}l@{}}\textbf{0.8163}\\\textbf{0.8641}\\\textbf{0.8680}\\\textbf{0.8775}\end{tabular}\\
		\hline
			Food101&\begin{tabular}[c]{@{}l@{}}12bits\\24bits\\32bits\\48bits\end{tabular}
			&\begin{tabular}[c]{@{}l@{}}0.1182\\0.1305\\0.1641\\0.2006\end{tabular}
			&\begin{tabular}[c]{@{}l@{}}0.2442\\0.3448\\0.3590\\0.3965\end{tabular}         
			&\begin{tabular}[c]{@{}l@{}}0.3564\\0.4093\\0.4289\\0.4881\end{tabular}
			&\begin{tabular}[c]{@{}l@{}}0.4563\\0.5548\\0.5639\\0.6419\end{tabular}
			&\begin{tabular}[c]{@{}l@{}}0.4644\\0.6687\\0.7427\\0.8213\end{tabular} 
			&\begin{tabular}[c]{@{}l@{}}0.5000\\0.7657\\0.8019\\0.8244\end{tabular}                                                                                  
			&\begin{tabular}[c]{@{}l@{}}\textbf{0.8306}\\\textbf{0.8585}\\\textbf{0.8635}\\\textbf{0.8642}\end{tabular}\\                                                                                       
		\hline
			NABirds&\begin{tabular}[c]{@{}l@{}}12bits\\24bits\\32bits\\48bits\end{tabular}
			&\begin{tabular}[c]{@{}l@{}}0.2170\\0.4080\\0.3610\\0.3200\end{tabular}
			&\begin{tabular}[c]{@{}l@{}}0.2340\\0.3290\\0.4520\\0.4970\end{tabular}         
			&\begin{tabular}[c]{@{}l@{}}0.2530\\0.8230\\0.1471\\0.2534\end{tabular}
			&\begin{tabular}[c]{@{}l@{}}0.5220\\0.1569\\0.2194\\0.3481\end{tabular}
			&\begin{tabular}[c]{@{}l@{}}0.0820\\0.1915\\0.2441\\0.3564\end{tabular} 
			&\begin{tabular}[c]{@{}l@{}}0.0812\\0.1944\\0.2826\\0.4115\end{tabular}                                                                                  
			&\begin{tabular}[c]{@{}l@{}}\textbf{0.6842}\\\textbf{0.7573}\\\textbf{0.7711}\\\textbf{0.7881}\end{tabular}\\                                                                                       
	 	\hline
	
		\hline

	\end{tabular}}
\end{table*}

\subsection{Hashing Code Generation}
After finishing the training phase, the binary hashing code $c_q$ is generated as:
\begin{equation}
\label{E12}
c_{q}=H'(f^{global}_q)=sign(W^hf^{global}_q)
\end{equation}

\section{EXPERIMENT}
	In this section, we show comprehensive experiments to verify the effectiveness of the proposed work. Firstly, we compare our model with the state-of-the-art methods on several publicly available fine-grained datasets. Then we explore the contribution of each proposed module. Following this, we perform additional experiments to demonstrate the effectiveness of the learnable hyperparameters and analyze the convergence of the network. Finally, we visualize the augmented images.

\subsection{Datasets and evaluation metrics}
	~~~~We evaluate the proposed method on several fine-grained datasets: 1) CUB-200-2011\cite{wah2011caltech} contains 11,788 images of 200 bird species, The dataset is divided into: the train set(5994 images) and the test set(5794 images), 2)FGVC Aircraft\cite{maji2013fine} contains 10,000 images of 100 aircraft variants, The dataset is divided into: the train set(6667 images) and the test set(3333 images), 3) Vegfru\cite{hou2017vegfru} includes 160,731 images covering 200 vegetable categories and 92 fruit categories, The dataset is divided into: the train set(29,200 images) and the test set(116,931 images), 4) Food101\cite{bossard2014food} contains 101 kinds of foods with 101,000 images, The dataset is divided into: the train set(75,750images) and the test set(25,250images), 5) NABirds\cite{van2015building} contains 48,562 images of North American birds with 555 sub-categories, The dataset is divided into: the train set(23,929 images) and the test set(24,633 images).

	To evaluate the retrieval performance, we adopted the Mean Average Precision (MAP) metric. 
\begin{equation}
\label{E14}
mAP=\frac{1}{n_q}\sum^{n_q}_{i=1}AP
\end{equation}
\begin{equation}
\label{E15}
AP=\frac{1}{n}\sum^L_{k=1}pos(k)
\end{equation}
\subsection{Comparative Methods}

	In this experiments, we employ six popular methods (e.g.  DPSH\cite{li2015feature}, HashNet\cite{cao2017hashnet}, ADSH\cite{jiang2018asymmetric}, ExchNet\cite{cui2020exchnet}, A2-Net\cite{wei20212} and Semicon\cite{shen2022semicon}), as benchmarks to prove the superiority of our proposed method.
	%Among them, the AA-NET method SEMICOM method is currently the latest ? 

\begin{figure*}[!t]
	\centering
	\includegraphics[width=7 in]{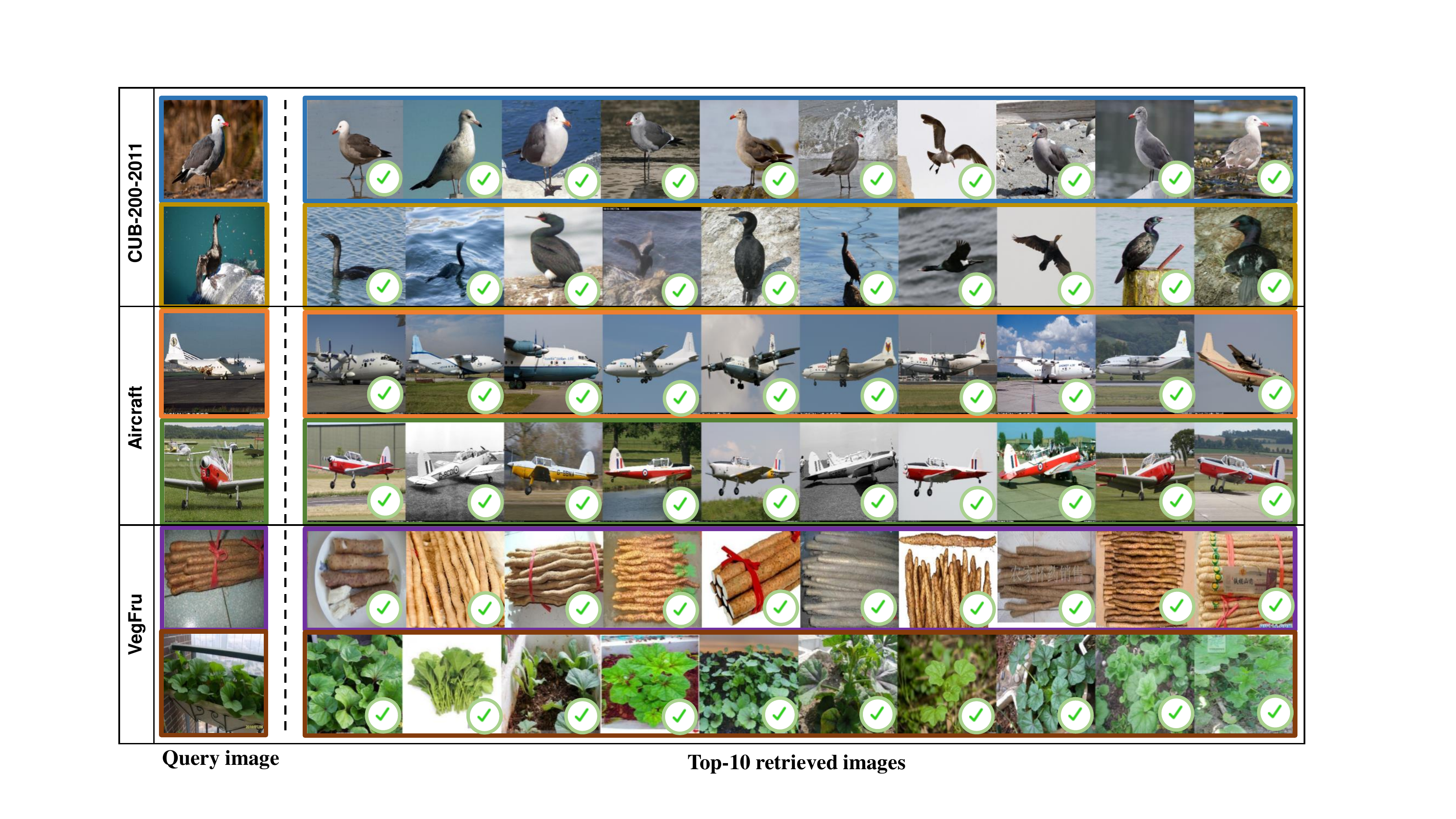}
	\caption{Examples of top-10 retrieved images on CUB200-2011, Aircraft and VegFru datasets}
	\label{fig3}
\end{figure*}
\subsection{Implementation Details and Experimental Settings}
	For a fair comparison, we implemented and conducted experiments using the same base model as the fine-grained baseline. Specifically, the experiments use the pre-trained Resnet50 as the backbone. throughout the experiments this work only requires the category information of the images and no additional annotation information such as bounding boxes.

	In the training phase, for pre-processing different types of datasets, this work crops a random region of the input image and then resizes it to 224×224 and flips it randomly horizontally.In the test phase, different types of datasets are adjusted to the input image size of 256 × 256, which is then centered and cropped to a size of 224 × 224. 

	This work uses a batch stochastic gradient descent algorithm (SGD) with a momentum of 0.9, a weight decay of 0.0003 used, a minimum batch size of 128, and a number of iterations of 150. For the FGVC-Aircraft dataset, we use an initial learning rate of 0.0035 and a mini-batch size of 32. For the VegFru dataset, we use an initial learning rate of 0.005 and a mini-batch size of 128. For the CUB-200-2011, Food101 and NABirds datasets, we use an initial learning rate of 0.008 and a mini-batch size of 128.

\subsection{Experimental Results for Hashing}
	In this subsection, we analyze the experimental results. For the five publicly available datasets mentioned in Section 4.1, we conducted experiments on these datasets using different hash bits, and the experimental results are shown in Table~\ref{tab1}, where the bolded ones represent the highest MAP values. We can find that our proposed model is very effective on all five fine-grained datasets. Our method achieves the best retrieval performance among all methods, both on traditional fine-grained datasets and on large-scale fine-grained datasets. In particular, on the NABirds dataset, with 48-bit hash codes, the MAP values obtained by our method far exceed the MAP values obtained by Semicon, the current best model, which shows the remarkable effectiveness of our method. In addition, on the VegFru dataset, our method shows a 10\% relative improvement over Semicon. On the Food101 dataset, our method has nearly 5\% relative improvement over Semicon.
	%And on the CUB200-2011 and datasets, our method also has \%-\% as well as \%-\% relative improvement over Semicon.

	Fig.~\ref{fig3} shows the returned results of the fine-grained retrieval system based on this method, where the first column of each row is the query image and the second to eleventh columns are the top ten most similar images returned by the system.

\subsection{Exploration Experiment}

\subsubsection{Cascaded Hierarchical Network Analysis}
Table~\ref{tab3} shows the experimental results. The results show that the best retrieval performance is achieved when we use the cascade network, indicating that our proposed cascade network is effective. Specifically, when generating 32bits hash codes, the MAP value is only 0.7831 when using only the top-level features, and the MAP value can be improved by about 4.7\% when cascading the features of other deep networks to generate hash codes. In particular, the best retrieval accuracy is achieved when cascading all the different deep network features. This demonstrates the advantage of using cascaded networks in fine-grained hashing.

\begin{table}[H]
	\caption{Contribution of proposed cascade hiearchical networks. }
	\label{tab3}
	\centering
	\begin{tabular}{|*{4}{c|}}
	\hline

	\hline
	\multicolumn{3}{|c|}{Configurations}&\textbf{CUB-200-2011}\\
     \cline{1-3}
		$x_{3}$ & $x_{2}$ & $x_{1}$&12bits 24bits 32bits  48bits\\
	\hline
		$\surd$& &              &0.0177 0.7653 0.7832 0.8216\\
		$\surd$&$\surd$&        &0.7709 0.8208 0.8284 0.8354\\
		$\surd$&       &$\surd$ &0.7669 0.8194 0.8258 0.8339\\
		$\surd$&$\surd$&$\surd$ &\textbf{0.7714 0.8219 0.8309 0.8392}\\
 	\hline

	\hline

	\end{tabular}
\end{table}

\subsubsection{Component Analysis of the Loss Function}
	In this section, we verify the effectiveness of the key components of the loss function in the proposed method. We decompose the loss function and conduct ablation experiments on CUB-200-2011. The loss function in training consists of two major key components, one is the feature representation learning component $L_{CLS}$ and the other is the hash code learning component $L_{HASH}$. The feature representation learning part $L_{CLS}$ is further divided into two parts: the fine-grained representation learning part $L_{cls\_org}$ of the original image and the fine-grained representation learning part $L_{cls\_aug}$ of the attention-based data augmented image.The experimental results are shown in Table~\ref{tab4}.

\begin{table}[H]
	\caption{Contribution of proposed components and their combinations. }
	\label{tab4}
	\centering
	\begin{tabular}{|*{4}{c|}}
	\hline

	\hline
	\multirow{2}*{$L_{HASH}$}& \multicolumn{2}{c|}{$L_{CLS}$}&\textbf{CUB-200-2011}\\ 
     \cline{2-3}
		& $L_{cls\_org}$ & $L_{cls\_aug}$&12bits 24bits 32bits  48bits\\
     \hline
		$\surd$& &              &0.7191 0.8049 0.8224 0.8304\\
		$\surd$&$\surd$&        &0.7710 0.8136 0.8285 0.8334\\
		$\surd$&       &$\surd$ &0.7511 0.8181 0.8270 0.8335\\
		$\surd$&$\surd$&$\surd$ &\textbf{0.7714 0.8219 0.8309 0.8392}\\
 	\hline

	\hline

	\end{tabular}
\end{table}

\subsubsection{Learnable Hyper-Parameters Analysis}
	In this subsections, we study the effect of the learnable hyper-parameters. The experiments are conducted on CUB-200-2011. Specifically, we compared the model retrieval performance in two cases, i.e., experiments using learnable parameters and experiments using fixed parameters.

     In the experiments using fixed parameters, we simply set {$\lambda$} as \{0.5, 0.8, 1.0\} and set {$\beta$} as \{0.8, 1.0\}. In the experiments using learnable parameters, we used 1 as the initial value for both {$\lambda$} and {$\beta$}.  Table~\ref{tab4} shows the retrieval results obtained in the CUB-200-2011 dataset using learnable parameters or using fixed parameters. The results show that the retrieval accuracy obtained with fixed parameters is generally not high, and the highest retrieval performance is achieved when the learnable parameters are used. This indicates that the learnable parameters are effective in the network.

	Fig.~\ref{fig3} shows the trend of dynamics a and b during the training of CUB-200-2011 dataset and VegFru dataset. As can be seen from the figure, both $\alpha$ and $\beta$ are dynamically updated while the network is being trained. On the CUB-200-2011 dataset, $\alpha$  eventually converges to 0.7 and $\beta$ eventually converges to 1.9. On the VegFru dataset, $\alpha$  eventually converges to 0.5 and $\beta$ eventually converges to 1.9. It can be seen from Equation 12 that the hash task accounts for a larger proportion of both tasks. It indicates that the hash task, which is the main task, dominates the training process.

\begin{table}[H]
	\caption{MAP in the testing set with fixed hyperparameters of loss functions on CUB-200-2011}
	\label{tab4}
	\centering
	\begin{tabular}{|*{4}{c|}}
	\hline

	\hline
	 \multirow{2}*{Method}&\multicolumn{2}{c|}{Parameter}&\textbf{CUB-200-2011}\\
	 \cline{2-3}
		    &$\alpha$&$\beta$&12bits 24bits 32bits  48bits \\
	 \hline
	    \multirow{4}*{Fixed}&1&0.8&0.7620 0.8046 0.8237 0.8262\\
	    	    &1  &1&0.7643 0.8096 0.8224 0.8328\\
	         &0.8&1&0.7703 0.8141 0.8200 0.8280\\
	         &0.5&1&0.7645 0.7973 0.8095 0.8174\\
 	\hline
        Dynamic & - & - & \textbf{0.7714 0.8219 0.8309 0.8392}\\

	\hline

	\end{tabular}
\end{table}

\begin{figure}[H]
	\centering
	\includegraphics[width=3.5 in]{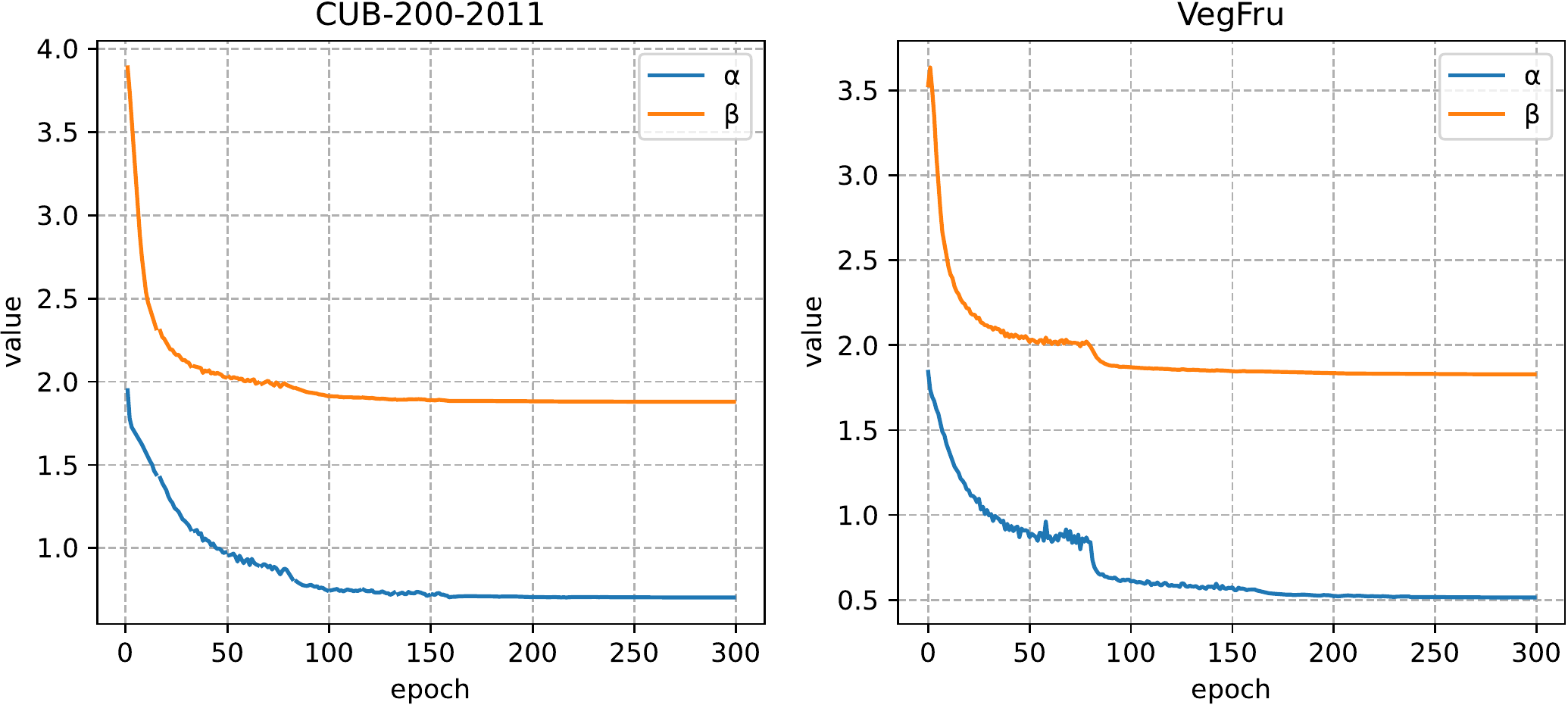}
	\caption{chart of the changing trend of dynamic hyper-parameters along with training on CUB-200-2011}
	\label{fig3}
\end{figure}

\subsubsection{Convergence Analysis }
	Fig.~\ref{fig4} shows the convergence curves of MAP values on the CUB-200-2011 dataset and the VegFru dataset, demonstrating the change of MAP values during the training of the network. It can be seen that on both data, the network starts to converge and the MAP values stabilize around 80 epoch.
\begin{figure}[H]
	\centering
	\includegraphics[width=3.5 in]{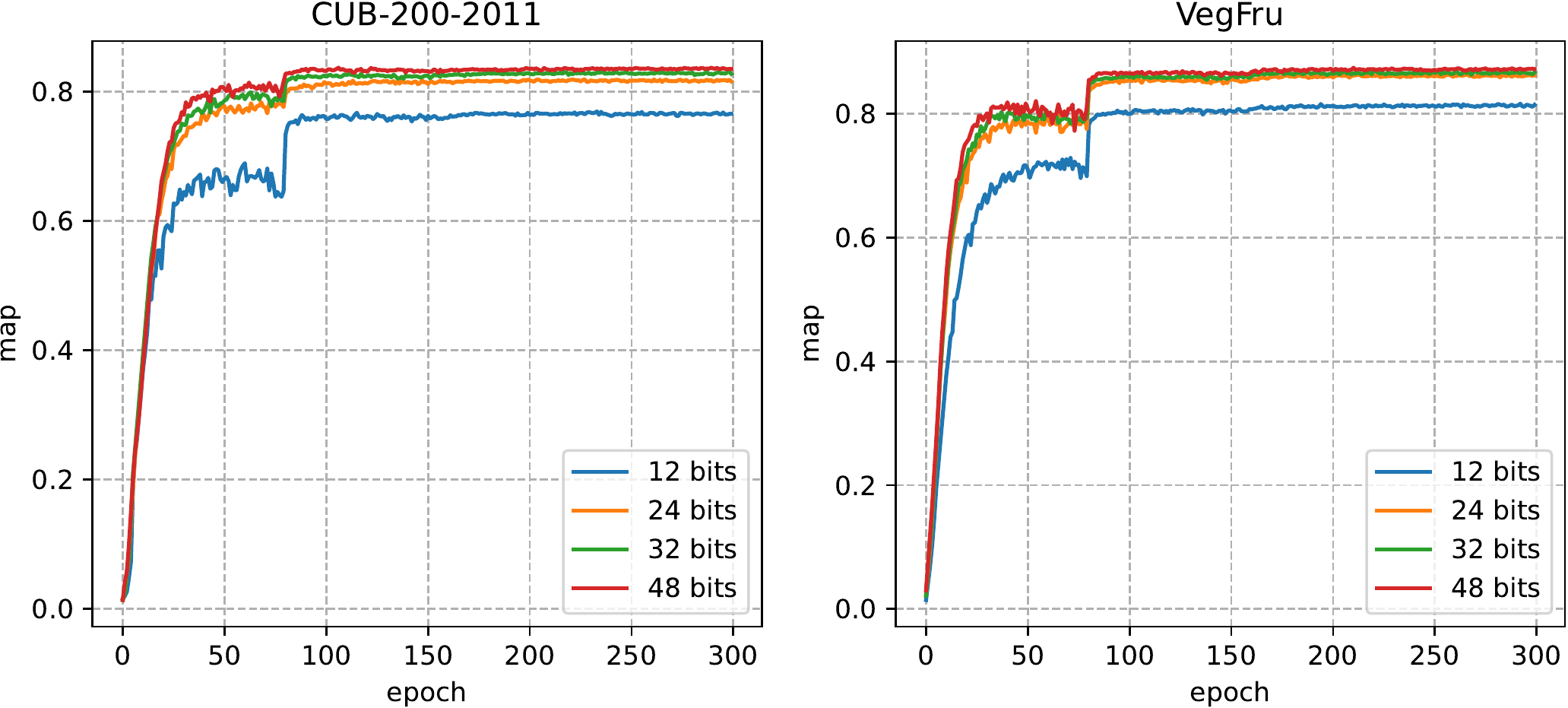}
	\caption{The convergence curves of the MAP values on CUB-200-2011 dataset and VegFru dataset.}
	\label{fig4}
\end{figure}

\subsubsection{Visualization of Augmented Data }
\begin{figure}[H]
	\centering
	\includegraphics[width=3.5 in]{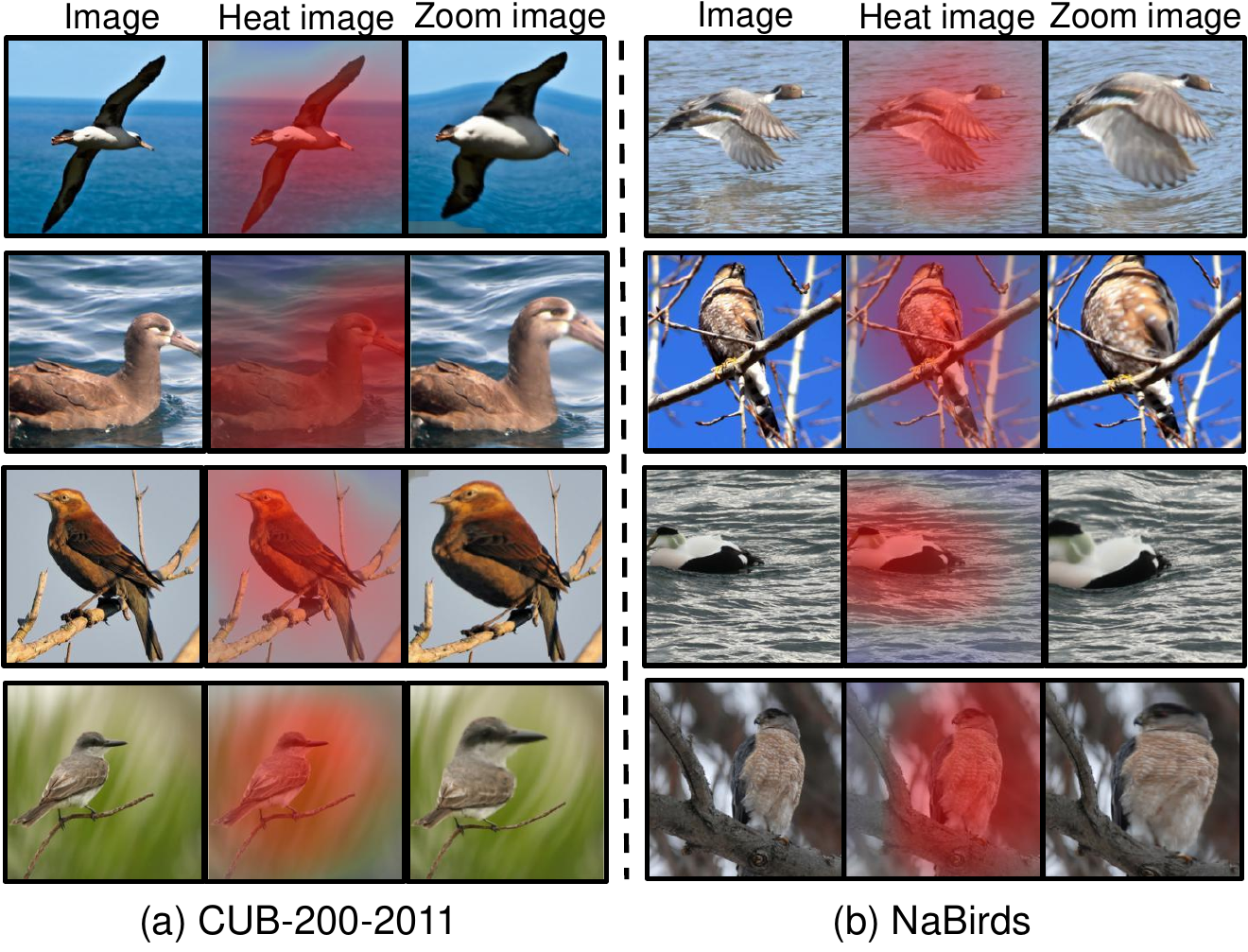}
	\caption{Examples of attention zooming on CUB-200-2011 dataset and NaBirds dataset}
	\label{fig_zoom}
\end{figure}

Fig.~\ref{fig_zoom} shows the augmented images obtained by attentional guidance in the CUB-200-2011 and NABirds datasets. Where the first column is the original image, the second column is the heat map of the attention region, and the third column is the attention-enhanced image. As seen from the attention heat map, the model can successfully focus on the target object and ignore the distracting background.
%%%%%%%%%%%%%%%%%%%%%%%%%%%%%%%%%%%%%%%%%%%%%%%%%%%%%%%%%%%%%%%%%%%%%%%%%%%%%%
\section{Conclusion}
In this paper, we propose a unified end-to-end trainable cascade network and introduce a novel approach to balance multitask learning. We experiment on many fine-grained image datasets, and the experimental results prove that our approach outperforms comparative algorithms and achieves the best performance. Based on this paper, subsequent work we consider these methods to solve classification, retrieval or other multi-task learning problems. We will also explore the contribution of self-attention mechanism to hashing, and further improve the accuracy of the attention mechanism for image object localization with a new attention module.

\ifCLASSOPTIONcompsoc
  \section*{Acknowledgments}
\else
  \section*{Acknowledgment}
\fi

This work was supported by the Open Research Fund from the Guangdong Provincial Key Laboratory of Big Data Computing, the Chinese University of Hong Kong, Shenzhen (No.B10120210117-OF08) and the Guangdong Province Ordinary Colleges and Universities Young Innovative Talents Project Grant (No. 2022KQNCX038).

\ifCLASSOPTIONcaptionsoff
  \newpage
\fi

%\begin{thebibliography}{1}
\bibliographystyle{IEEEtran}
\bibliography{ref2}

\begin{thebibliography}{10}
\providecommand{\url}[1]{#1}
\csname url@samestyle\endcsname
\providecommand{\newblock}{\relax}
\providecommand{\bibinfo}[2]{#2}
\providecommand{\BIBentrySTDinterwordspacing}{\spaceskip=0pt\relax}
\providecommand{\BIBentryALTinterwordstretchfactor}{4}
\providecommand{\BIBentryALTinterwordspacing}{\spaceskip=\fontdimen2\font plus
\BIBentryALTinterwordstretchfactor\fontdimen3\font minus
  \fontdimen4\font\relax}
\providecommand{\BIBforeignlanguage}[2]{{%
\expandafter\ifx\csname l@#1\endcsname\relax
\typeout{** WARNING: IEEEtran.bst: No hyphenation pattern has been}%
\typeout{** loaded for the language `#1'. Using the pattern for}%
\typeout{** the default language instead.}%
\else
\language=\csname l@#1\endcsname
\fi
#2}}
\providecommand{\BIBdecl}{\relax}
\BIBdecl

\bibitem{wei2019rpc}
X.-S. Wei, Q.~Cui, L.~Yang, P.~Wang, and L.~Liu, ``Rpc: A large-scale retail
  product checkout dataset,'' \emph{arXiv preprint arXiv:1901.07249}, 2019.

\bibitem{hou2017vegfru}
S.~Hou, Y.~Feng, and Z.~Wang, ``Vegfru: A domain-specific dataset for
  fine-grained visual categorization,'' in \emph{Proceedings of the IEEE
  International Conference on Computer Vision}, 2017, pp. 541--549.

\bibitem{wang2017survey}
J.~Wang, T.~Zhang, N.~Sebe, H.~T. Shen \emph{et~al.}, ``A survey on learning to
  hash,'' \emph{IEEE transactions on pattern analysis and machine
  intelligence}, vol.~40, no.~4, pp. 769--790, 2017.

\bibitem{lin2016learning}
K.~Lin, J.~Lu, C.-S. Chen, and J.~Zhou, ``Learning compact binary descriptors
  with unsupervised deep neural networks,'' in \emph{Proceedings of the IEEE
  conference on computer vision and pattern recognition}, 2016, pp. 1183--1192.

\bibitem{xie2015fine}
L.~Xie, J.~Wang, B.~Zhang, and Q.~Tian, ``Fine-grained image search,''
  \emph{IEEE Transactions on multimedia}, vol.~17, no.~5, pp. 636--647, 2015.

\bibitem{he2015delving}
K.~He, X.~Zhang, S.~Ren, and J.~Sun, ``Delving deep into rectifiers: Surpassing
  human-level performance on imagenet classification,'' in \emph{Proceedings of
  the IEEE international conference on computer vision}, 2015, pp. 1026--1034.

\bibitem{xiong2016achieving}
W.~Xiong, J.~Droppo, X.~Huang, F.~Seide, M.~Seltzer, A.~Stolcke, D.~Yu, and
  G.~Zweig, ``Achieving human parity in conversational speech recognition,''
  \emph{arXiv preprint arXiv:1610.05256}, 2016.

\bibitem{lecun2015deep}
Y.~LeCun, Y.~Bengio, and G.~Hinton, ``Deep learning,'' \emph{nature}, vol. 521,
  no. 7553, pp. 436--444, 2015.

\bibitem{gordo2016deep}
A.~Gordo, J.~Almaz{\'a}n, J.~Revaud, and D.~Larlus, ``Deep image retrieval:
  Learning global representations for image search,'' in \emph{European
  conference on computer vision}.\hskip 1em plus 0.5em minus 0.4em\relax
  Springer, 2016, pp. 241--257.

\bibitem{xu2015augmenting}
Z.~Xu, S.~Huang, Y.~Zhang, and D.~Tao, ``Augmenting strong supervision using
  web data for fine-grained categorization,'' in \emph{Proceedings of the IEEE
  international conference on computer vision}, 2015, pp. 2524--2532.

\bibitem{krause2016unreasonable}
J.~Krause, B.~Sapp, A.~Howard, H.~Zhou, A.~Toshev, T.~Duerig, J.~Philbin, and
  L.~Fei-Fei, ``The unreasonable effectiveness of noisy data for fine-grained
  recognition,'' in \emph{Computer Vision--ECCV 2016: 14th European Conference,
  Amsterdam, The Netherlands, October 11-14, 2016, Proceedings, Part III
  14}.\hskip 1em plus 0.5em minus 0.4em\relax Springer, 2016, pp. 301--320.

\bibitem{zhang2020web}
C.~Zhang, Y.~Yao, J.~Zhang, J.~Chen, P.~Huang, J.~Zhang, and Z.~Tang,
  ``Web-supervised network for fine-grained visual classification,'' in
  \emph{2020 IEEE International Conference on Multimedia and Expo
  (ICME)}.\hskip 1em plus 0.5em minus 0.4em\relax IEEE, 2020, pp. 1--6.

\bibitem{zhang2020web2}
C.~Zhang, Y.~Yao, H.~Liu, G.-S. Xie, X.~Shu, T.~Zhou, Z.~Zhang, F.~Shen, and
  Z.~Tang, ``Web-supervised network with softly update-drop training for
  fine-grained visual classification,'' in \emph{Proceedings of the AAAI
  Conference on Artificial Intelligence}, vol.~34, no.~07, 2020, pp.
  12\,781--12\,788.

\bibitem{zhang2014part}
N.~Zhang, J.~Donahue, R.~Girshick, and T.~Darrell, ``Part-based r-cnns for
  fine-grained category detection,'' in \emph{Computer Vision--ECCV 2014: 13th
  European Conference, Zurich, Switzerland, September 6-12, 2014, Proceedings,
  Part I 13}.\hskip 1em plus 0.5em minus 0.4em\relax Springer, 2014, pp.
  834--849.

\bibitem{huang2016part}
S.~Huang, Z.~Xu, D.~Tao, and Y.~Zhang, ``Part-stacked cnn for fine-grained
  visual categorization,'' in \emph{Proceedings of the IEEE conference on
  computer vision and pattern recognition}, 2016, pp. 1173--1182.

\bibitem{lin2017bilinear}
T.-Y. Lin, A.~RoyChowdhury, and S.~Maji, ``Bilinear convolutional neural
  networks for fine-grained visual recognition,'' \emph{IEEE transactions on
  pattern analysis and machine intelligence}, vol.~40, no.~6, pp. 1309--1322,
  2017.

\bibitem{gao2016compact}
Y.~Gao, O.~Beijbom, N.~Zhang, and T.~Darrell, ``Compact bilinear pooling,'' in
  \emph{Proceedings of the IEEE conference on computer vision and pattern
  recognition}, 2016, pp. 317--326.

\bibitem{yu2018hierarchical}
C.~Yu, X.~Zhao, Q.~Zheng, P.~Zhang, and X.~You, ``Hierarchical bilinear pooling
  for fine-grained visual recognition,'' in \emph{Proceedings of the European
  conference on computer vision (ECCV)}, 2018, pp. 574--589.

\bibitem{cai2017higher}
S.~Cai, W.~Zuo, and L.~Zhang, ``Higher-order integration of hierarchical
  convolutional activations for fine-grained visual categorization,'' in
  \emph{Proceedings of the IEEE international conference on computer vision},
  2017, pp. 511--520.

\bibitem{bell2015learning}
S.~Bell and K.~Bala, ``Learning visual similarity for product design with
  convolutional neural networks,'' \emph{ACM transactions on graphics (TOG)},
  vol.~34, no.~4, pp. 1--10, 2015.

\bibitem{zhang2016embedding}
X.~Zhang, F.~Zhou, Y.~Lin, and S.~Zhang, ``Embedding label structures for
  fine-grained feature representation,'' in \emph{Proceedings of the IEEE
  Conference on Computer Vision and Pattern Recognition}, 2016, pp. 1114--1123.

\bibitem{huang2016local}
C.~Huang, C.~C. Loy, and X.~Tang, ``Local similarity-aware deep feature
  embedding,'' \emph{Advances in neural information processing systems},
  vol.~29, 2016.

\bibitem{wang2019multi}
X.~Wang, X.~Han, W.~Huang, D.~Dong, and M.~R. Scott, ``Multi-similarity loss
  with general pair weighting for deep metric learning,'' in \emph{Proceedings
  of the IEEE/CVF conference on computer vision and pattern recognition}, 2019,
  pp. 5022--5030.

\bibitem{jin2020deep}
S.~Jin, H.~Yao, X.~Sun, S.~Zhou, L.~Zhang, and X.~Hua, ``Deep saliency hashing
  for fine-grained retrieval,'' \emph{IEEE Transactions on Image Processing},
  vol.~29, pp. 5336--5351, 2020.

\bibitem{zheng2018centralized}
X.~Zheng, R.~Ji, X.~Sun, Y.~Wu, F.~Huang, and Y.~Yang, ``Centralized ranking
  loss with weakly supervised localization for fine-grained object retrieval.''
  in \emph{IJCAI}, 2018, pp. 1226--1233.

\bibitem{zheng2019towards}
X.~Zheng, R.~Ji, X.~Sun, B.~Zhang, Y.~Wu, and F.~Huang, ``Towards optimal fine
  grained retrieval via decorrelated centralized loss with normalize-scale
  layer,'' in \emph{Proceedings of the AAAI conference on artificial
  intelligence}, vol.~33, no.~01, 2019, pp. 9291--9298.

\bibitem{qian2019softtriple}
Q.~Qian, L.~Shang, B.~Sun, J.~Hu, H.~Li, and R.~Jin, ``Softtriple loss: Deep
  metric learning without triplet sampling,'' in \emph{Proceedings of the
  IEEE/CVF International Conference on Computer Vision}, 2019, pp. 6450--6458.

\bibitem{zeng2020fine}
X.~Zeng, Y.~Zhang, X.~Wang, K.~Chen, D.~Li, and W.~Yang, ``Fine-grained image
  retrieval via piecewise cross entropy loss,'' \emph{Image and Vision
  Computing}, vol.~93, p. 103820, 2020.

\bibitem{kulis2009kernelized}
B.~Kulis and K.~Grauman, ``Kernelized locality-sensitive hashing for scalable
  image search,'' in \emph{2009 IEEE 12th international conference on computer
  vision}.\hskip 1em plus 0.5em minus 0.4em\relax IEEE, 2009, pp. 2130--2137.

\bibitem{datar2004locality}
M.~Datar, N.~Immorlica, P.~Indyk, and V.~S. Mirrokni, ``Locality-sensitive
  hashing scheme based on p-stable distributions,'' in \emph{Proceedings of the
  twentieth annual symposium on Computational geometry}, 2004, pp. 253--262.

\bibitem{kong2012isotropic}
W.~Kong and W.-J. Li, ``Isotropic hashing,'' \emph{Advances in neural
  information processing systems}, vol.~25, 2012.

\bibitem{li2015feature}
W.-J. Li, S.~Wang, and W.-C. Kang, ``Feature learning based deep supervised
  hashing with pairwise labels,'' \emph{arXiv preprint arXiv:1511.03855}, 2015.

\bibitem{liu2016deep}
H.~Liu, R.~Wang, S.~Shan, and X.~Chen, ``Deep supervised hashing for fast image
  retrieval,'' in \emph{Proceedings of the IEEE conference on computer vision
  and pattern recognition}, 2016, pp. 2064--2072.

\bibitem{su2017deep}
S.~Su, G.~Chen, X.~Cheng, and R.~Bi, ``Deep supervised hashing with nonlinear
  projections.'' in \emph{IJCAI}, 2017, pp. 2786--2792.

\bibitem{jiang2018asymmetric}
Q.-Y. Jiang and W.-J. Li, ``Asymmetric deep supervised hashing,'' in
  \emph{Proceedings of the AAAI conference on artificial intelligence},
  vol.~32, no.~1, 2018.

\bibitem{cui2020exchnet}
Q.~Cui, Q.-Y. Jiang, X.-S. Wei, W.-J. Li, and O.~Yoshie, ``Exchnet: A unified
  hashing network for large-scale fine-grained image retrieval,'' in
  \emph{European Conference on Computer Vision}.\hskip 1em plus 0.5em minus
  0.4em\relax Springer, 2020, pp. 189--205.

\bibitem{wei20212}
X.-S. Wei, Y.~Shen, X.~Sun, H.-J. Ye, and J.~Yang, ``A $^2$-net: Learning
  attribute-aware hash codes for large-scale fine-grained image retrieval,''
  \emph{Advances in Neural Information Processing Systems}, vol.~34, pp.
  5720--5730, 2021.

\bibitem{chen2022fine}
Z.-D. Chen, X.~Luo, Y.~Wang, S.~Guo, and X.-S. Xu, ``Fine-grained hashing with
  double filtering,'' \emph{IEEE Transactions on Image Processing}, vol.~31,
  pp. 1671--1683, 2022.

\bibitem{simonyan2014very}
K.~Simonyan and A.~Zisserman, ``Very deep convolutional networks for
  large-scale image recognition,'' \emph{arXiv preprint arXiv:1409.1556}, 2014.

\bibitem{he2016deep}
K.~He, X.~Zhang, S.~Ren, and J.~Sun, ``Deep residual learning for image
  recognition,'' in \emph{Proceedings of the IEEE conference on computer vision
  and pattern recognition}, 2016, pp. 770--778.

\bibitem{huang2017densely}
G.~Huang, Z.~Liu, L.~Van Der~Maaten, and K.~Q. Weinberger, ``Densely connected
  convolutional networks,'' in \emph{Proceedings of the IEEE conference on
  computer vision and pattern recognition}, 2017, pp. 4700--4708.

\bibitem{cao2017hashnet}
Z.~Cao, M.~Long, J.~Wang, and P.~S. Yu, ``Hashnet: Deep learning to hash by
  continuation,'' in \emph{Proceedings of the IEEE international conference on
  computer vision}, 2017, pp. 5608--5617.

\bibitem{shen2022semicon}
Y.~Shen, X.~Sun, X.-S. Wei, Q.-Y. Jiang, and J.~Yang, ``Semicon: A
  learning-to-hash solution for large-scale fine-grained image retrieval,'' in
  \emph{Computer Vision--ECCV 2022: 17th European Conference, Tel Aviv, Israel,
  October 23--27, 2022, Proceedings, Part XIV}.\hskip 1em plus 0.5em minus
  0.4em\relax Springer, 2022, pp. 531--548.

\bibitem{recasens2018learning}
A.~Recasens, P.~Kellnhofer, S.~Stent, W.~Matusik, and A.~Torralba, ``Learning
  to zoom: a saliency-based sampling layer for neural networks,'' in
  \emph{Proceedings of the European Conference on Computer Vision (ECCV)},
  2018, pp. 51--66.

\bibitem{wah2011caltech}
C.~Wah, S.~Branson, P.~Welinder, P.~Perona, and S.~Belongie, ``The caltech-ucsd
  birds-200-2011 dataset,'' 2011.

\bibitem{maji2013fine}
S.~Maji, E.~Rahtu, J.~Kannala, M.~Blaschko, and A.~Vedaldi, ``Fine-grained
  visual classification of aircraft,'' \emph{arXiv preprint arXiv:1306.5151},
  2013.

\bibitem{bossard2014food}
L.~Bossard, M.~Guillaumin, and L.~V. Gool, ``Food-101--mining discriminative
  components with random forests,'' in \emph{European conference on computer
  vision}.\hskip 1em plus 0.5em minus 0.4em\relax Springer, 2014, pp. 446--461.

\bibitem{van2015building}
G.~Van~Horn, S.~Branson, R.~Farrell, S.~Haber, J.~Barry, P.~Ipeirotis,
  P.~Perona, and S.~Belongie, ``Building a bird recognition app and large scale
  dataset with citizen scientists: The fine print in fine-grained dataset
  collection,'' in \emph{Proceedings of the IEEE Conference on Computer Vision
  and Pattern Recognition}, 2015, pp. 595--604.

\end{thebibliography}
%\end{thebibliography}

\begin{IEEEbiography}[{\includegraphics[width=1in,height=1.25in,clip,keepaspectratio]{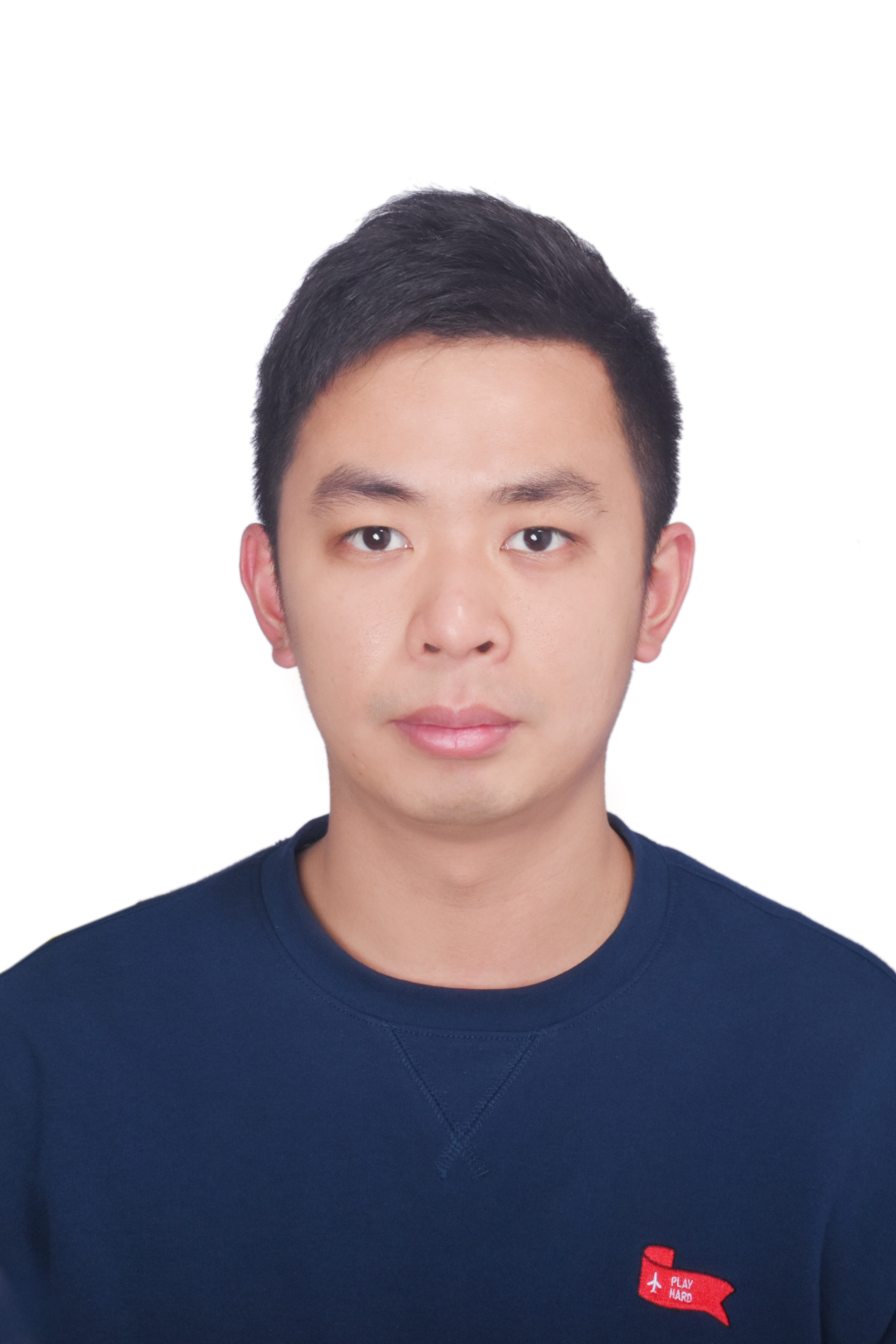}}]{Xianxian Zeng}
received the B.Eng. and Ph.D degree from the School of Automation, Guangdong University of Technology, Guangzhou, China, in 2015 and 2020, respectively. He is currently a Lecturer with the Sehool Computer Science, Guangdong Polytechnic Normal University. His research interests include computer vision, pattern recognition and machine leaning.
\end{IEEEbiography}

\begin{IEEEbiography}[{\includegraphics[width=1in,height=1.25in,clip,keepaspectratio]{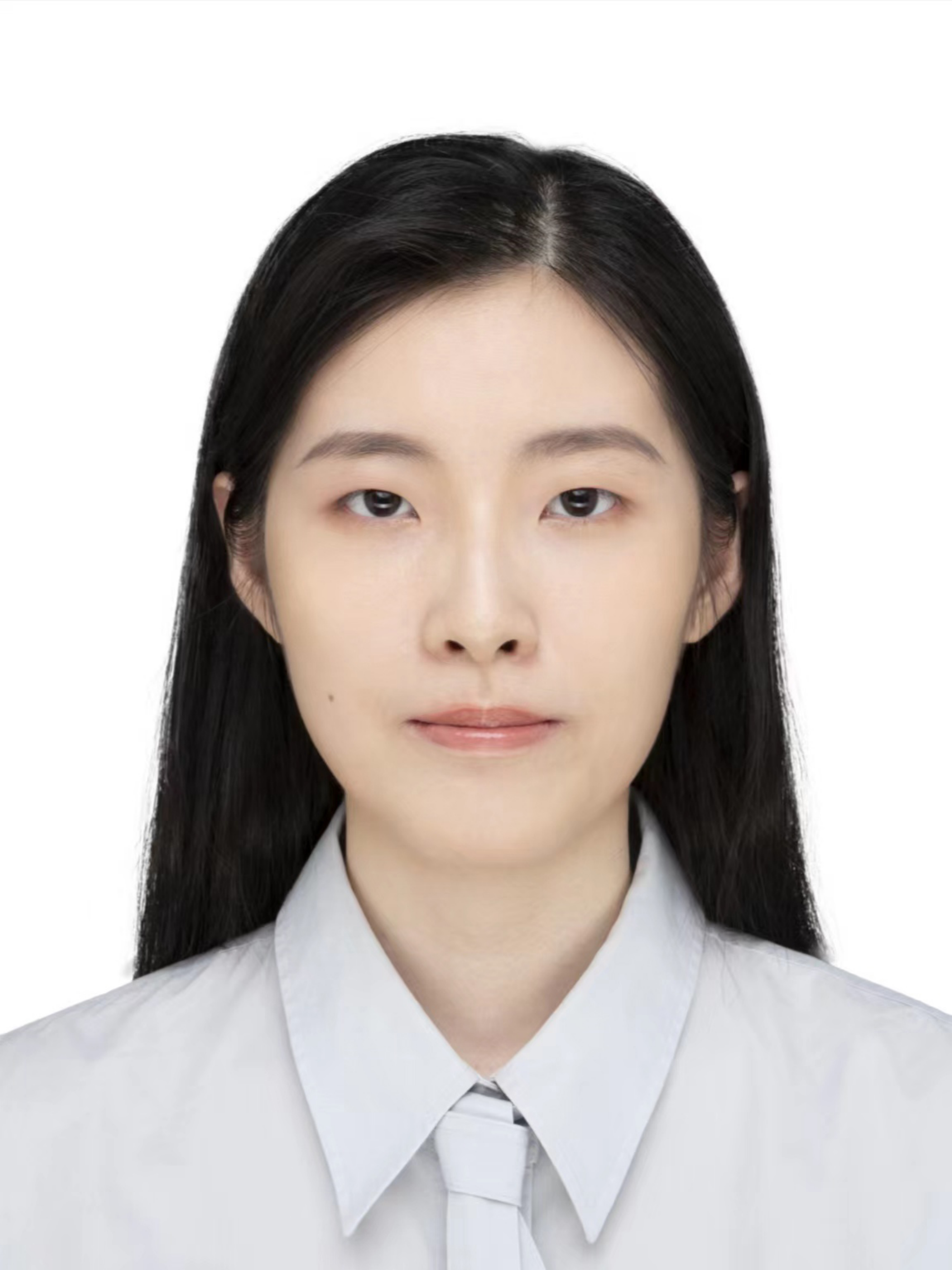}}]{Yanjun Zheng}
received the bachelor degree in computer science from Guangdong Polytechnic Normal University, China,
in 2022. Her research interests include deep learning, representation learning, fine-grained analysis.
\end{IEEEbiography}

%\newpage

%\begin{IEEEbiographynophoto}{ CCC}

\end{document}